\documentclass{article}

\usepackage{microtype}
\usepackage{graphicx}
\usepackage{subfigure}
\usepackage{booktabs} 

\usepackage{hyperref}

\usepackage[accepted]{icml2025}

\usepackage{amsmath}
\usepackage{amssymb}
\usepackage{mathtools}
\usepackage{amsthm}

\usepackage[capitalize,noabbrev]{cleveref}

\theoremstyle{plain}

\theoremstyle{definition}

\theoremstyle{remark}

\usepackage[textsize=tiny]{todonotes}

\icmltitlerunning{Maintaining MTEB: Towards Long Term Usability and Reproducibility of Embedding Benchmarks}

\begin{document}

\twocolumn[
\icmltitle{Maintaining MTEB: Towards Long Term Usability \\and Reproducibility of Embedding Benchmarks}



\icmlsetsymbol{equal}{*}

\begin{icmlauthorlist}
\icmlauthor{Isaac Chung}{equal,zen}
\icmlauthor{Imene Kerboua}{equal,esker,lyon}
\icmlauthor{Márton Kardos}{aarhus}
\icmlauthor{Roman Solomatin}{itmo}
\icmlauthor{Kenneth Enevoldsen}{aarhus}
\end{icmlauthorlist}

\icmlaffiliation{esker}{Esker}
\icmlaffiliation{lyon}{INSA Lyon, LIRIS}
\icmlaffiliation{itmo}{ITMO University}
\icmlaffiliation{zen}{Zendesk}
\icmlaffiliation{aarhus}{Aarhus University}

\icmlcorrespondingauthor{Isaac Chung}{chungisaac1217@gmail.com}
\icmlcorrespondingauthor{Imene Kerboua}{imene.kerboua@insa-lyon.fr}

\icmlkeywords{Machine Learning, ICML}

\vskip 0.3in
]



\printAffiliationsAndNotice{\icmlEqualContribution} 

\begin{abstract}

The Massive Text Embedding Benchmark (MTEB) has become a standard evaluation platform for text embedding models. While previous work has established the core benchmark methodology, this paper focuses on the engineering aspects that ensure MTEB's continued reproducibility and extensibility. We present our approach to maintaining robust continuous integration pipelines that validate dataset integrity, automate test execution, and assess benchmark results' generalizability. We detail the design choices that collectively enhance reproducibility and usability. Furthermore, we discuss our strategies for handling community contributions and extending the benchmark with new tasks and datasets. These engineering practices have been instrumental in scaling MTEB to become more comprehensive while maintaining quality and, ultimately, relevance to the field. Our experiences offer valuable insights for benchmark maintainers facing similar challenges in ensuring reproducibility and usability in machine learning evaluation frameworks. 

The MTEB repository is available at: 
\url{https://github.com/embeddings-benchmark/mteb}

\end{abstract}
\section{Introduction}

Benchmarks are critical in guiding machine learning progress, particularly in the rapidly evolving field of embeddings and representation learning. MTEB \cite{muennighoff2022mteb} has emerged as a comprehensive evaluation framework for embedding models across diverse tasks and languages. 
While initially designed as a standalone benchmark, MTEB has evolved into a versatile evaluation ecosystem. This natural evolution has been driven by sustained interest and engagement from the research community, integrating numerous extensions and specialized variants: language-specific benchmarks like C-MTEB \cite{cmteb2024}, MTEB-French \cite{ciancone2024mtebfrench}, and German Text Embedding Clustering Benchmark \cite{wehrli2024germantextembeddingclustering}; regional benchmarks such as SEB \cite{enevoldsen2024scandinavian}; multilingual expansions through MMTEB \cite{enevoldsen2025mmteb} covering over 250 languages; domain-specific adaptations like ChemTEB \cite{kasmaee2025chemteb}; and cross-modal evaluations like MIEB \cite{xiao2025mieb}.

However, as the benchmark grows in scope and adoption, maintaining its sustainability and effectiveness presents significant engineering challenges that have received limited attention in the literature. These challenges span several dimensions: \textbf{reproducibility} — ensuring results can be consistently replicated given a model; \textbf{usability} — evaluating and presenting comparisons effectively and efficiently; \textbf{extensibility} — accommodating new tasks, languages, and modalities without disrupting existing functionality; \textbf{integrity} — ensuring results reflect genuine generalization, free from training data contamination, and \textbf{relevance} — addressing current research challenges as the field evolves. The literature has emphasized benchmark relevance, but often neglects the practical infrastructure required to maintain a benchmark's utility over time.


This paper explores the technical design choices underpinning MTEB's sustainability and extensibility. We focus on the software engineering practices implemented to \textbf{1) ensure dataset quality}, \textbf{2) assess benchmark zero-shot levels}
, and \textbf{3) facilitate community engagement and expansion}. Through concrete case studies, we demonstrate how our approach addresses real-world challenges in benchmark maintenance, from multilingual expansion to contamination detection and community contribution scaling.

By sharing our experiences in maintaining MTEB, we aim to highlight often overlooked aspects of benchmark development that are essential for long-term usability. The engineering approaches described here address common challenges in benchmark maintenance and can inform similar efforts across the machine learning community, ultimately contributing to more reproducible and trustworthy evaluation standards. 

\section{The MTEB Framework}
Maintaining a large-scale benchmark repository like MTEB requires robust infrastructure that balances reproducibility and flexibility to accommodate diverse and novel models and evaluation scenarios. Rather than focusing solely on implementation details, we highlight the design principles that enable MTEB's sustainable growth and innovation potential.

\subsection{System Architecture}
\begin{figure*}[ht]
\centering
\includegraphics[width=\linewidth]{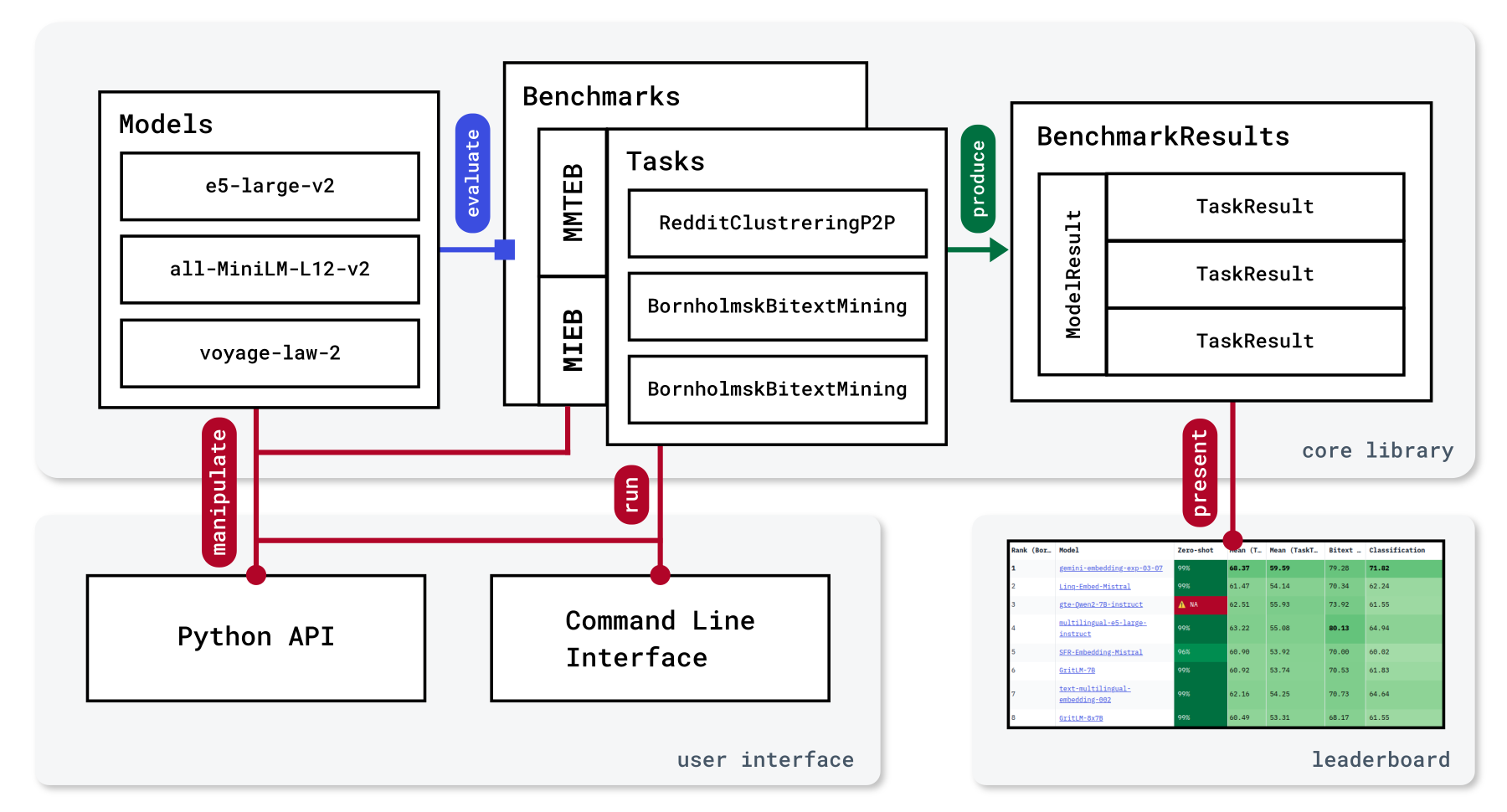}
\caption{MTEB is designed with a modular architecture that separates concerns into distinct components. Users can select which model, task, or benchmark to run via the Python API or command line interface. The produced results can then be submitted to the results repository, which is read by the leaderboard.}
\label{fig:system-arch}
\end{figure*}

MTEB consists of a set of code repositories and a leaderboard. \autoref{fig:system-arch} shows at a high level the relationship between models, tasks, and result aggregation:
\textbf{Standardized Model interfaces} for embedding generation, allowing evaluation of diverse architectures from sentence transformers to instruction-tuned LLMs with minimal adaptation;
\textbf{Task definitions} encapsulate evaluation logic for different paradigms (classification, clustering, retrieval, etc.) while remaining agnostic to specific datasets or models;
\textbf{Dataset handlers} manage data loading, preprocessing, and validation while maintaining consistent interfaces across languages and domains;
\textbf{Result processors} standardize output formats and compute metrics uniformly across tasks.





To automate validations and releases, we adopt standard CI/CD practice in the MTEB repository, which is detailed in \autoref{appdx:tech}. 

\subsection{Leaderboard}

In addition to the evaluation package, MTEB features an open leaderboard that displays model performance across the benchmark, enabling businesses or users to select suitable models for their needs and supporting researchers in driving progress in the field. For a visualization of the leaderboard, see \autoref{fig:mteb-lb} in \autoref{appdx:lb}. 

Submissions of results to the leaderboard are often made by the broader community. These can be made by submitting a pull request to the results repository.

\section{Ensuring Benchmark Reproducibility}

While MTEB's modular architecture provides the foundation for its functionality, specific reproducibility measures are essential to ensure the benchmark's validity over time. The primary challenges we identified revolve around ambiguity in the evaluation workflow: which version of a dataset or model is being used, which version of the code is executing the evaluation, and whether saved results can be reliably loaded and trusted.   

\subsection{Multi-level Versioning System}
A key challenge in benchmark maintenance is managing changes to various components while preserving result comparability. Our versioning system addresses this through: \textbf{Task versioning}: Tasks undergo updates when evaluation protocols are changed, e.g. using stratified sampling. Each version maintains its own evaluation protocol.
\textbf{Dataset versioning}: Each dataset references a specific revision from its source (typically Hugging Face), ensuring identical data across evaluation runs. Versions are updated when issues like mislabeled examples, or duplicates are discovered. 
\textbf{Model versioning}: Models reference specific checkpoints or API versions to maintain consistency, as model behavior can change significantly between releases.
\textbf{Code versioning}: The MTEB package itself follows semantic versioning, with clear compatibility boundaries.

\subsection{Reproducible Evaluation Environments}

Certain evaluation methods introduce potential variability that must be controlled. For example, clustering tasks using K-means and classification tasks using linear probing require hyperparameter search procedures that can produce different results based on initialization. To address this, we provide consistent seeds for deterministic execution of evaluation tasks.
Additionally, recognizing the substantial computational resources required for benchmark evaluation, we added $CO_2$ emissions logging with the \verb|codecarbon| library \footnote{\url{https://codecarbon.io/}}. This helps users quantify and report the environmental impact of their evaluation runs, promoting transparency in the computational costs of benchmarking. 

\subsection{Community Verifiable Results}

The MTEB leaderboard is designed to maximize transparency and verifiability. Submissions require:
\textbf{Reference implementation}: Each model must have a well-documented reference implementation. For open-source models, this includes direct links to code repositories and model weights. For proprietary models, implementations may rely on API calls, though this introduces a trust assumption that the API consistently serves the same model version.
\textbf{Peer review process}: Submitted results undergo community review via pull requests to the results repository, where maintainers and community members can verify methodology and question unusual patterns.
\textbf{Version-specific tracking}: Results explicitly record which versions of MTEB, datasets, and models were used, enabling precise reproduction of evaluation conditions.

This comprehensive approach ensures that anyone can reproduce a benchmark result by running the specified dataset version against the specified model version using a specific MTEB version, yielding identical results across different machines and environments.




These reproducibility measures reflect lessons learned from past benchmark failures (See \autoref{sec:case-study-2}) and help MTEB serve as a robust foundation for fair and consistent evaluation of embedding models.
\section{Design for Extensibility}
MTEB is an evaluation framework and should first and foremost be reproducible and thus stable. To achieve this, MTEB utilizes a modular design intended to allow contributors to extend any component without modifying others, enabling growth while preserving backward compatibility and result comparability.

Building MTEB to be extensible also encourages innovation. Any newly added custom models, tasks, and benchmarks can also be reproduced using the same underlying framework. 

Unlike many existing benchmarks that enforce an often restrictive interface for model evaluation \cite{nielsen2024encoder, li2024coir}, MTEB provides models with rich contextual information likely to be available at inference time, including task type, targeted language, or task-specific prompts, enabling researchers to explore diverse model architectures and prompting strategies. 
This flexibility allows models to be evaluated more realistically and creatively, promoting progress in representation learning.
By incorporating recent models, tasks, and benchmarks, MTEB remains relevant to the field organically. 


\section{Case Studies}

The sustainability of a benchmark depends not only on infrastructure and reproducibility, but also on community engagement and demonstrated utility. 
Community involvement not only strengthens the reliability of the benchmark, but also helps ensure its ongoing maintenance, relevance, and representativeness, particularly for low-resource communities \cite{singh2024ayadatasetopenaccesscollection}.
Overall, over 170 contributors participated in the development of the project. Their collective efforts have been instrumental in shaping MTEB into a multilingual, multi-domain, multimodal, community-driven benchmark.
This section presents key challenges we encountered while maintaining MTEB and the solutions we implemented, offering practical insights for other benchmark maintainers.

\subsection{Case Study 1: Assessing zero-shot levels.}  


\begin{figure}[h]
\centering
\includegraphics[width=\linewidth]{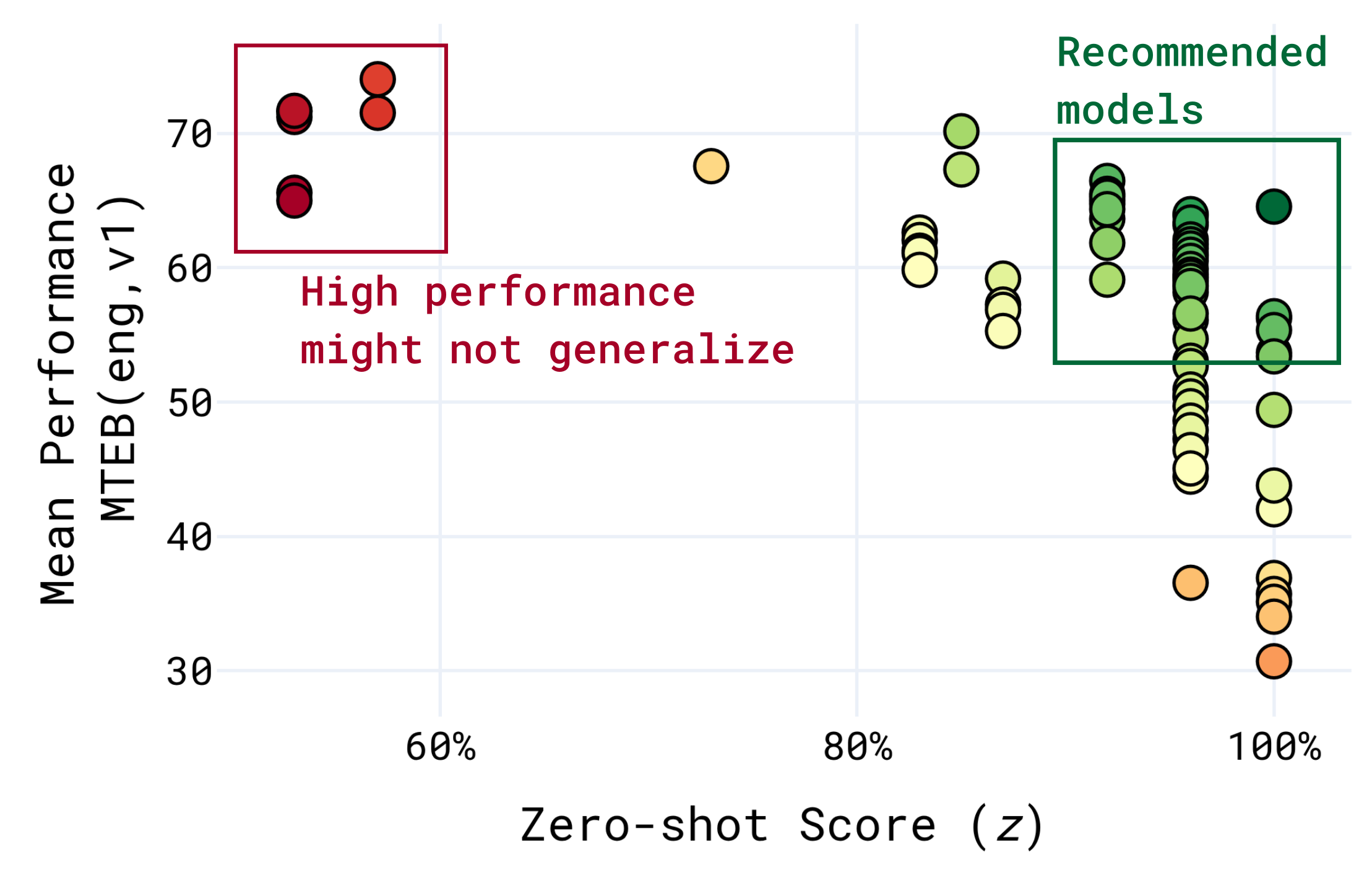}
\caption{Models' mean performance against their zero-shot score on the legacy English MTEB. 
The highest ranking models achieve their scores by training on benchmark tasks, even though models with lower scores might generalize better to out-of-distribution environments.
}
\label{fig:zero-shot}
\end{figure}

A significant challenge in embedding benchmarks is determining whether models genuinely demonstrate out-of-distribution generalization. Unlike traditional data leakage (where test examples appear in training data) \cite{magar-schwartz-2022-data,choi2025contaminatedbenchmarkquantifyingdataset}, MTEB faces a more nuanced challenge: models being trained on datasets with similar distributions to benchmark tasks, particularly when using training splits from the same source as benchmark evaluation tasks. 
To address this, we implemented a transparency approach that: 1) encourages model contributors to disclose training datasets, 2) computes a \textit{zero-shot score} quantifying distributional overlap:


\[
z = 1 - \frac{n_{\text{train}}}{n_{\text{total}}}
\]

where \( z \) is the zero-shot score, \( n_{\text{train}} \) is the number of benchmark datasets the model was trained on, and \( n_{\text{total}} \) is the total number of benchmark datasets.

This metric offers a coarse estimate of how ``in-domain'' or ``out-of-domain'' a model is with respect to the benchmark. For example, the \texttt{e5-mistral-7b-instruct} model \cite{wang-etal-2024-improving-text} has a 95\% zero-shot score on MTEB (English, v2), indicating that it was trained on only $\sim$5\% of the benchmark's training splits, i.e. a 5\% \textit{leak}. Despite this, the model performs strongly across the board, suggesting strong generalization to unseen tasks.

Our initial implementation filtered the leaderboard to show only models with 100\% zero-shot scores or unknown training data. This approach proved too restrictive, as community members noted it penalized transparency by hiding otherwise strong models that honestly disclosed their training data. Following this feedback, we developed a more granular approach that preserves transparency while providing users with critical context for interpreting benchmark scores.
This case study demonstrates how benchmarks must balance methodological rigor with practical usability, and how community feedback can drive improvements in benchmark design.
Detailed key exchanges on this topic can be found in \autoref{appdx:zs}.

\subsection{Case Study 2: Reproducing Reported Results.}
\label{sec:case-study-2}
The MTEB leaderboard recently transitioned from relying on self-reported results in Hugging Face model cards to a centralized repository of verified results. This transition revealed several challenges in reproducing reported model performance: 1) embedding models can use prefixes (E5 \cite{wang2024multilinguale5textembeddings} using \texttt{query:} and \texttt{passage:} as query and passage prompts for retrieval), 2) prompts can differ per task-type (Nomic models), 3) prompts can be added to both the query and the passage during retrieval (E5-mistral\cite{wang2023improving}) or only to queries (NV Embed), 4) some models do not normalize embeddings (Nomic ModernBERT), 5) models can have custom parameters during encoding (jina-v3 \cite{sturua2024jinaembeddingsv3multilingualembeddingstask} \texttt{task\_type} argument that loads LoRAs during inference), and 6) model can have additional stages (CDE \cite{morris2024contextualdocumentembeddings} have different stages for encoding to store embeddings). 

To ensure accurate evaluation across this diverse landscape of embedding model architectures, we implemented additional features that accommodate each of these requirements. 
For instance, implementing prefix support allowed us to reproduce the reported performance of BGE models, which had previously shown significant performance degradation when evaluated.
The improved reproducibility increased community confidence in the leaderboard rankings.
The full details of the related pull requests are in \autoref{appdx:repro}.


\section{Limitations}
\label{appdx:limitations}

The framework does not currently address issues in bias \cite{rakivnenko2024biastextembeddingmodels,Cao_2025}. Coverage in other diverse domains like arts, culture, health \cite{cao2024recentadvancestextembedding} is also limiting.  

MTEB has also received criticism from model makers for its loose adherence to semantic versioning and backwards compatibility.
These issues mainly came about as a result of the large refactoring effort that was required for multilingual expansion.
While these refactoring effort did benefit the community and the package at large,
breaking changes were at times introduced in minor/patch releases,
which frequent users of the package found confusing and frustrating.
We have since then put more emphasis on maintaining backward compatibility with older versions of the library to avoid further inconvenience.

Since the initial launch of MTEB, the package has maintained a simple documentation structure, initially in the README and as it expanded across multiple hyperlinked markdown files. Although this was initially simple to maintain, the code base has grown in size and complexity and is now more prone to code-documentation inconsistencies. We plan to incorporate documentation within the docstrings and generate docs automatically in future releases (v2.0.0 or above).

\section{Conclusion}
MTEB’s transition from a standalone benchmark to a scalable evaluation ecosystem has hinged on robust design choices. We present our approach to maintaining MTEB towards better reproducibility, maintainability, and seamless integration of new tasks and modalities. These infrastructure choices have enabled broad community engagement and sustained growth without compromising quality. 
Our experience offers practical guidance for designing and maintaining benchmarks, contributing to more reproducible, reliable, and globally representative machine learning research.






\section*{Impact Statement}
MTEB promotes equitable progress in language technology by enabling rigorous, scalable evaluation across high- and low-resource languages.  



\section*{Acknowledgements}
We thank the reviewers for their insightful feedback and all contributors to the MTEB library.


\bibliography{main}
\bibliographystyle{icml2025}

\newpage
\appendix

\section{Full Descriptions of Technical Infrastructure}
\label{appdx:tech}

\subsection{Continuous Integration}

Our continuous integration (CI) pipeline validates both code and datasets, a step often neglected in other benchmarks \cite{thakur2021beir,laion2025clipbenchmark}.


\subsection{Dataset and Model Validation}
Any pull request that adds or modifies datasets triggers automated checks, including: 1) format consistency, 2) metadata completeness, 3) metadata field validation, and 4) and dataset availability on Hugging Face. Similarly, any new or modified model metadata is automatically checked for completeness and model loading. Both types of validations heavily rely on Pydantic\footnote{https://github.com/pydantic/pydantic}. These help catch malformed examples and pre-processing inconsistencies early.

Each dataset and model has their respective Pydantic model that specifies the mandatory fields to capture crucial information for reproducibility. Specifically, each dataset has \texttt{TaskMetadata} and each model has \texttt{ModelMeta}. For example, Table \ref{tab:modelmeta-parameters} shows all of the \texttt{ModelMeta} parameters and their descriptions as of version 1.38.4.

\begin{table*}[t]
\centering
\resizebox{\linewidth}{!}{
\begin{tabular}{r|l|l}
\hline
\textbf{Parameter} & \textbf{Type} & \textbf{Description} \\
\hline
\texttt{loader} & Callable or None & Function to load the model; defaults to SentenceTransformer if None. \\
\texttt{name} & str & Name of the model, ideally in the format ``organization/model\_name''. \\
\texttt{n\_parameters} & int or None & Number of parameters in the model, e.g., 7\_000\_000. \\
\texttt{memory\_usage\_mb} & float or None & Memory usage of the model in MB. \\
\texttt{max\_tokens} & int or None & Maximum tokens the model can handle. \\
\texttt{embed\_dim} & int & Dimension of embeddings produced. \\
\texttt{revision} & str or None & Model revision identifier. \\
\texttt{release\_date} & str & Date the revision was released. \\
\texttt{license} & str & License under which the model is released. \\
\texttt{open\_weights} & bool & Whether model weights are open source. \\
\texttt{public\_training\_code} & str or None & URL to training code if publicly available. \\
\texttt{public\_training\_data} & str or None & URL to training data if publicly available. \\
\texttt{similarity\_fn\_name} & str & Distance metric used by the model. \\
\texttt{framework} & list of str & List of frameworks used, e.g., \texttt{["Sentence Transformers", "PyTorch"]}. \\
\texttt{reference} & str & URL to the model's page (e.g., HuggingFace). \\
\texttt{languages} & list of str & Languages supported, e.g., \texttt{["eng-Latn"]}. \\
\texttt{use\_instructions} & bool & Whether the model uses instruction-based formatting. \\
\texttt{training\_datasets} & dict & Datasets used in training, e.g., \texttt{\{"ArguAna": ["test"]\}}. \\
\texttt{adapted\_from} & str or None & Base model this one was adapted from. \\
\texttt{superseded\_by} & str or None & Model that supersedes this one. \\
\texttt{is\_cross\_encoder} & bool & Whether the model functions as a cross-encoder. \\
\texttt{modalities} & list of str & Modalities supported, default is \texttt{["text"]}. \\
\hline
\end{tabular}
}
\caption{Parameters of the \texttt{ModelMeta} class}
\label{tab:modelmeta-parameters}
\end{table*}

\subsection{Pull Request Checks}
Contributions to the codebase are subject to an automated validation workflow comprising three sequential stages: 1) Linting and formatting, 2) Execution of unit and integration tests, 3) Verification of any required documentation updates. 

By delegating routine checks to this pipeline, the burden on maintainers is reduced and the project remains receptive to contributions at scale.

\subsubsection{Linting}
All incoming changes are analyzed with ruff\footnote{\url{https://github.com/astral-sh/ruff}} according to a shared configuration that enforces coding conventions (e.g., import ordering, modernized Python syntax) and detects common errors.

\subsubsection{Testing}
To ensure reproducibility, the test suite is executed on both Linux and Windows platforms, with Linux jobs covering Python versions 3.9 through 3.12. This cross‐platform strategy has been instrumental in uncovering environment‐specific defects during development.

\textbf{Mock infrastructure} We employ mock tasks—each defined by two to three illustrative examples—and synthetic models that emit random embeddings. These mocks enable rapid feedback by exercising task evaluators and model‐interaction code without incurring the overhead of full model instantiation.

\textbf{Parameter validation} Tests assert that all mandatory parameters are propagated correctly from evaluator modules to model classes under a variety of configurations, and that prompting routines apply inputs as intended.

\textbf{Data integration} Certain tests dynamically retrieve datasets from the Hugging Face Hub to validate end‐to‐end data loading and preprocessing.

\textbf{Reliability enhancement} We integrate the \texttt{pytest-rerunfailures} plugin to automatically retry tests that fail due to transient connection errors, thereby improving overall suite stability.

\subsection{Versioning, Releases, and Automations}

\textbf{Versioning:} We follow a SemVer-like versioning strategy adapted to benchmarks: 1) major versions indicate code-incompatible changes, 2) minor versions add tasks or datasets without affecting existing results, and 3) patch versions fix bugs or update documentation.

\textbf{Changelog and Release Automation:} Releases automatically generate changelogs with: 1) contributor lists and 2) change descriptions. Version-tagged releases trigger: 1) PyPI publishing, 2) documentation deployment, and 3) release note preparation. This ensures reproducible and traceable versions.

\textbf{Automated Artifacts:} To standardize result reporting, we support automated generation of: 1) Markdown tables of benchmark and tasks with descriptions and citations, 2) descriptive statistics of datasets, and 3) language coverage of the entire repository.
These tools improve result clarity and reduce manual reporting errors. Latex tables can be generated at will for academic publications and are not automated. 

\textbf{Leaderboard:} The MTEB leaderboard is refreshed and rebuilt daily. The previous workflow involved scraping the huggingface model cards for self-reported scores. The current workflow requires active submissions to the results repository. 
\section{Leaderboard}
\label{appdx:lb}

\begin{figure*}[h]
\centering
\resizebox{\linewidth}{!}{\includegraphics[width=\linewidth]{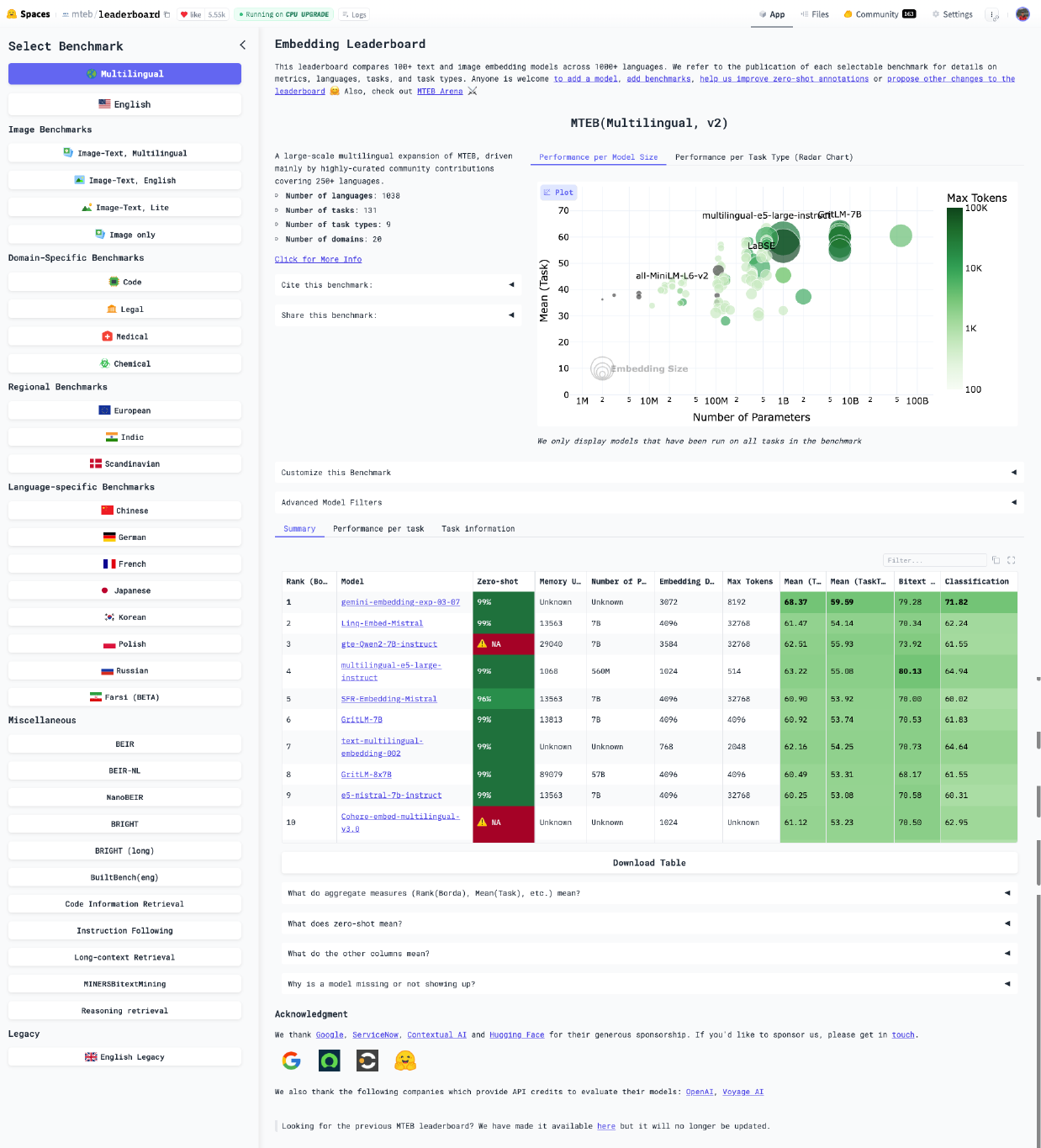}}
\caption{The MTEB Leaderboard offers an expandable collection of embedding benchmarks, with its default set as Multilingual (MMTEB, \cite{enevoldsen2025mmteb}). For every benchmark, a model performance per model size graph is shown, with an option to show model performance per task type in a separate tab. Models are ranked by Borda Count by default. The MTEB leaderboard also offers options to customize a benchmark by filtering on fields such as task types, languages, and domain. The left sidebar contains highlighted benchmarks by groups, while all benchmarks are available in the code. }
\label{fig:mteb-lb}
\end{figure*}

\autoref{fig:mteb-lb} shows the MTEB leaderboard in light mode.

\section{Model Zero-Shot Score}
\label{appdx:zs}
\newpage

Key discussions on zero-shot scores are in this section:  
\begin{itemize}
    \item Discussion on the definition of zero-shot: \url{https://github.com/embeddings-benchmark/mteb/discussions/2351}
    \item Community feedback on the initial version and filtering of zero-shot models on the leaderboard: \url{https://github.com/embeddings-benchmark/mteb/discussions/2119}
\end{itemize} 
\section{Reproducing Reported Results}
\label{appdx:repro}

We provide example pull requests on each of the takeaways during this period on benchmark reproducibility: 
\begin{enumerate}
  \item Embedding models can use prefixes (e5, bge models): \url{https://github.com/embeddings-benchmark/mteb/issues/1912}
  \item Prompts can differ per task-type (Nomic models): \url{https://github.com/embeddings-benchmark/mteb/pull/1685}
  \item Prompts can be added to both the query and the passage during retrieval (E5-mistral\cite{wang2023improving}) or only to queries (Nvembed): \url{https://github.com/embeddings-benchmark/mteb/pull/1436}
  \item Some models do not normalize embeddings (Nomic ModernBERT):  \url{https://github.com/embeddings-benchmark/mteb/pull/1684}
  \item Models can have custom parameters during encoding (jina-v3\cite{sturua2024jinaembeddingsv3multilingualembeddingstask} \texttt{task\_type} argument that loads LoRAs during inference): \url{https://github.com/embeddings-benchmark/mteb/pull/1319} 
  \item Models can have additional stages (CDE \cite{morris2024contextualdocumentembeddings} have different stages for encoding to store embeddings): \url{https://github.com/embeddings-benchmark/mteb/pull/2076} 
\end{enumerate}


\end{document}